\documentclass[runningheads]{llncs}
\usepackage{graphicx}
\usepackage{cite}
\usepackage{amsmath,amssymb,amsfonts}
\usepackage{algorithmic}
\usepackage{graphicx}
\usepackage{textcomp}
\usepackage{adjustbox}
\usepackage{multirow}
\usepackage{bbold}
\usepackage{siunitx}
\usepackage{hyperref}       
\usepackage{xcolor}
\usepackage{sidecap}
\hypersetup{
    colorlinks,
    linkcolor={blue!50!black},
    citecolor={blue!50!black},
    urlcolor={blue!50!black}
}

\begin{document}
\title{Spatially Varying Label Smoothing:\\Capturing Uncertainty from Expert Annotations}
%
\titlerunning{Spatially Varying Label Smoothing}

\author{Mobarakol Islam \and Ben Glocker}


\institute{BioMedIA Group, Department of Computing, Imperial College London, UK\\
\email{(m.islam20, b.glocker)@imperial.ac.uk}\\
}

\maketitle              
\begin{abstract}
The task of image segmentation is inherently noisy due to ambiguities regarding the exact location of boundaries between anatomical structures. We argue that this information can be extracted from the expert annotations at no extra cost, and when integrated into state-of-the-art neural networks, it can lead to improved calibration between soft probabilistic predictions and the underlying uncertainty. We built upon label smoothing (LS) where a network is trained on `blurred' versions of the ground truth labels which has been shown to be effective for calibrating output predictions. However, LS is not taking the local structure into account and results in overly smoothed predictions with low confidence even for non-ambiguous regions. Here, we propose Spatially Varying Label Smoothing (SVLS), a soft labeling technique that captures the structural uncertainty in semantic segmentation. SVLS also naturally lends itself to incorporate inter-rater uncertainty when multiple labelmaps are available. The proposed approach is extensively validated on four clinical segmentation tasks with different imaging modalities, number of classes and single and multi-rater expert annotations. The results demonstrate that SVLS, despite its simplicity, obtains superior boundary prediction with improved uncertainty and model calibration.

\end{abstract}
\section{Introduction}
\label{sec:introduction}
Understanding the prediction uncertainty is crucial in critical decision-making tasks like medical diagnosis. Despite impressive performance of deep neural networks, the output predictions of such models are often poorly calibrated and over-confident \cite{guo2017calibration, muller2019does, pereyra2017regularizing, laves2019well}. There is evidence that strategies such as label smoothing (LS) \cite{szegedy2016rethinking, muller2019does, pereyra2017regularizing} and temperature scaling \cite{guo2017calibration, kull2019beyond, laves2019well} are useful for calibration and uncertainty quantification for independent class prediction tasks such as image classification. Semantic segmentation, however, is a highly structured problem where pixel-wise class predictions intrinsically depend on the spatial relationship of neighboring objects. In medical images boundaries between anatomical regions are often ambiguous or ill-defined due to image noise, low contrast or the presence of pathology which should be taking into account during model training. Disagreement between experts annotators is also quite common and poses a challenge to the definition of `ground truth' in terms of hard segmentation labels. This requires mechanisms for aggregating multiple annotations in a reliable and sensible way that is able to capture the underlying uncertainty. Previous work fuses labels by using different techniques such as majority voting \cite{iglesias2015multi} or STAPLE \cite{warfield2004simultaneous}. However, these techniques do not carry the inter-rater variability through to the model predictions. Other approaches are integrating uncertainty in segmentation directly as part of the model's ability to make probabilistic predictions \cite{jungo2018effect, baumgartner2019phiseg, monteiro2020stochastic}. Nair et al. \cite{nair2020exploring} use Monte Carlo dropout \cite{gal2015dropout} to estimate uncertainty for multiple sclerosis lesion segmentation. Jungo et al. \cite{jungo2020analyzing} analyse the uncertainty and calibration of brain tumor segmentation using U-Net-like \cite{ronneberger2015u} architectures. These techniques are complementary to what we propose here and could be considered in a combined approach.

The utility of model uncertainty is directly related to model calibration which indicates how well the confidence estimates of a probabilistic classifier reflects the true error distribution. A predictive model should not only be accurate but also well-calibrated such that the provided confidence (or uncertainty) becomes useful in practice. Various calibration techniques \cite{guo2017calibration} such as Platt scaling \cite{platt1999probabilistic}, temperature scaling with dropout variational inference \cite{laves2019well}, non-parametric calibration \cite{wenger2020non}, Dirichlet calibration \cite{kull2019beyond}, and also label smoothing \cite{muller2019does, pereyra2017regularizing} have been proposed to improve calibration of output predictions. 

Label smoothing is an intriguingly simple technique which can both incorporate information about ambiguities in expert labels and improve model calibration. LS flattens the hard labels by assigning a uniform distribution over all other classes, which prevents over-confidence in the assigned label during training. LS has been shown to improve performance in deep learning models applied to different applications, including classification \cite{szegedy2016rethinking}, recognition \cite{vaswani2017attention} and language modeling \cite{chorowski2016towards}. Recently, M{\"u}ller et al. \cite{muller2019does} show that LS can significantly improve model calibration with a class separating effect on the learned feature representations.

In semantic segmentation, however, the benefit of LS is less clear. The notion of structured prediction via pixel-wise classification with underlying spatial dependencies, limits LS ability to perform as well as in image classification. This is because LS flattens the training labels with a uniform distribution without considering spatial consistency. Moreover, both convolution and pooling techniques in a fully convolutional network are performed on local spatially-varying patches. Therefore, simply generating pixel-wise soft labels that ignore the underlying structure may be insufficient for capturing uncertainty in semantic segmentation, as demonstrated empirically in our experiments.

We propose a simple yet effective mechanism, \textit{Spatially Varying Label Smoothing} (SVLS), for incorporating structural label uncertainty by capturing ambiguity about object boundaries in expert segmentation maps. SVLS does not require any additional annotations, and extracts uncertainty information directly from available expert annotation. Training a state-of-the-art neural network with SVLS target labels results in improved model calibration where uncertainty estimates for the target predictions do more faithfully represent segmentation accuracy. SVLS also naturally lends itself to the case where multiple expert annotations are available. To demonstrate the effectiveness of SVLS, we present results on four different clinical tasks including MRI segmentation of brain tumours and prostate zones, and CT segmentation of kidney tumours and lung nodules.

\begin{figure*}[!t]
    \centering
    \includegraphics[width=1\textwidth]{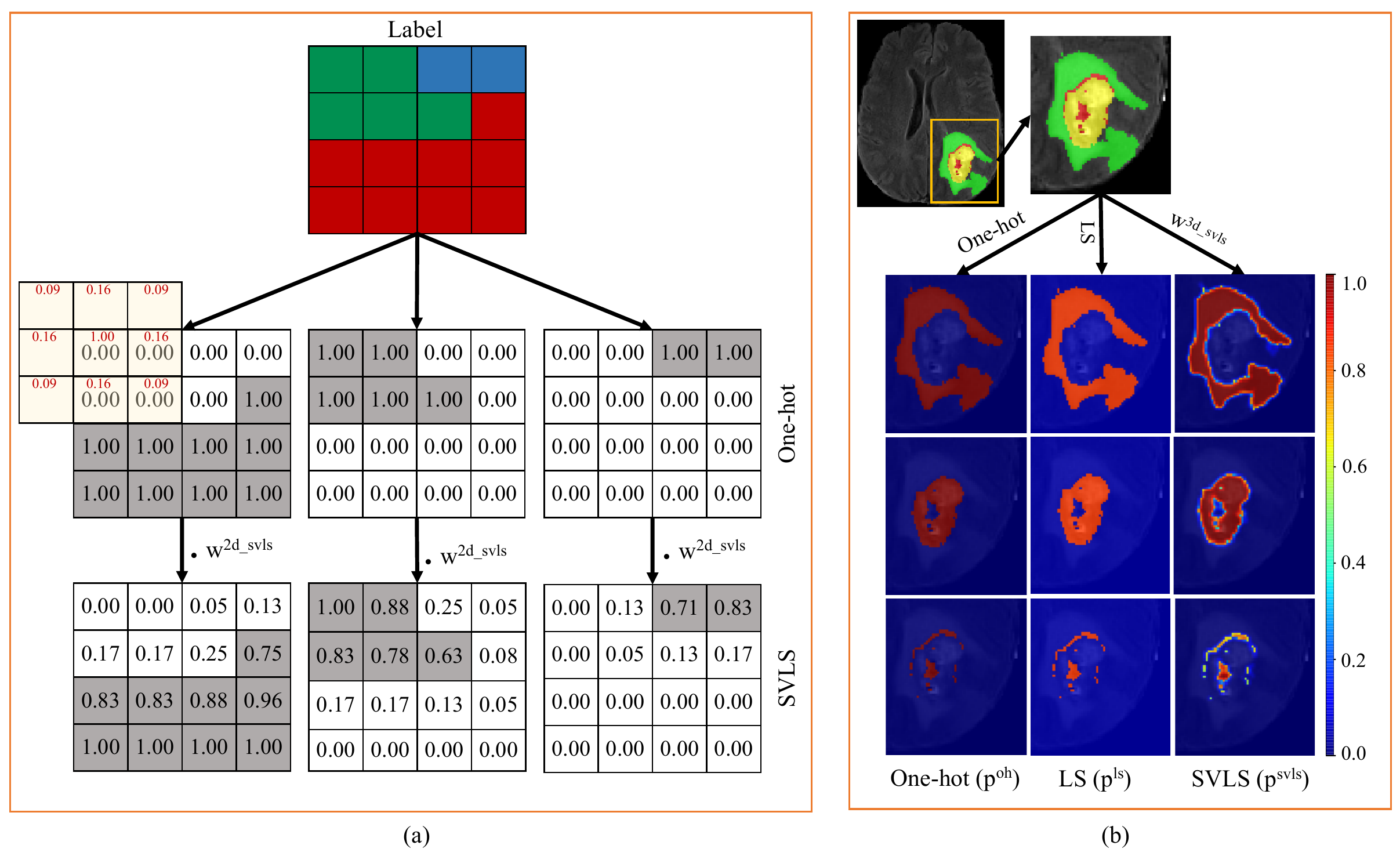}
    \caption{SVLS label generation and comparison with one-hot and LS. (a) A toy example of calculating the SVLS soft labels from expert segmentations. `Replicate' padding of the size 1 is for the calculation; (b) A visualization comparing the target labelmaps of one-hot, LS and SVLS for a multi-class brain tumour segmentation. One-hot and LS ignore the underlying spatial structure, whereas SVLS captures boundary uncertainty while preserving high confidence inside structures.}
    \label{fig:weights_labels}
\end{figure*}

\section{Spatially Varying Label Smoothing}

In semantic segmentation, a commonly used loss function is cross-entropy, $CE = - \sum_{c=1}^N{p_c^{oh}} \log(\hat{p_c})$, where $p_c^{oh}$ is the target label probability (or one-hot encoding), $\hat{p_c}$ is the predicted likelihood probability, with $N$ being the number of classes. In label smoothing (LS), the target probability is uniformly downscaled with a weight factor $\alpha$ to generate soft labels from the one-hot encoding, as
\begin{eqnarray}
\label{eq:LS}
p_c^{ls} = p_c^{oh}(1-\alpha)+\frac{\alpha}{N}
\end{eqnarray}
Simply replacing the original targets $p_c^{oh}$ with LS targets $p_c^{ls}$ in the CE loss allows us to train a neural network for image segmentation with soft target labels.

The LS approach originally proposed for image classification distributes the label uncertainty uniformly across classes and independently of the underlying spatial structure. For semantic segmentation, which is commonly formulated as dense pixel-wise classification, it is important to consider that the individual predictions are actually not independent, and neither are the class labels. The likelihood of the co-occurrence of different classes is spatially varying, in particular, near object boundaries which typically exhibit higher uncertainty \cite{wickstrom2020uncertainty, rottmann2019uncertainty, vu2019tunet, boutry2019using}. In order to take these aspects into account, we introduce Spatially Varying Label Smoothing (SVLS) which is an adaptation of LS for the task of semantic segmentation. SVLS determines the probability of the target class based on neighboring pixels. This is achieved by designing a weight matrix with a Gaussian kernel which is applied across the one-hot encoded expert labelmaps to obtain soft class probabilities. The SVLS weight matrix, $w^{svls}$, is obtained by taking the weights from a discrete spatial Gaussian kernel $G_{ND}(\vec{x};\sigma) = \frac{1}{(\sqrt {2\pi \sigma^2} )^N} e^{ - \frac{|\vec{x}|^2}{2\sigma ^2}}$, with $\sigma$ set to 1. The center weight is replaced by the sum of the surrounding weights, and we then normalize all weights by dividing with the center weight such that the center weight is equal to one. The motivation of this weight matrix is based on the intuition of giving an equal contribution to the central label and all surrounding labels combined. In most cases, the central class label would define which class is assigned the highest label probability in the SVLS soft labels. However, note that the weight matrix has an interesting property, which concerns cases where the central pixel has a class label in the expert annotation that is different from another label that may be present in all surrounding pixels. Such isolated, central class labels can be safely assumed to correspond to label noise, and in such cases the weights on the surrounding labels would result in an equal class probability.

\begin{figure}[!t]
    \centering
    \includegraphics[width=1\textwidth]{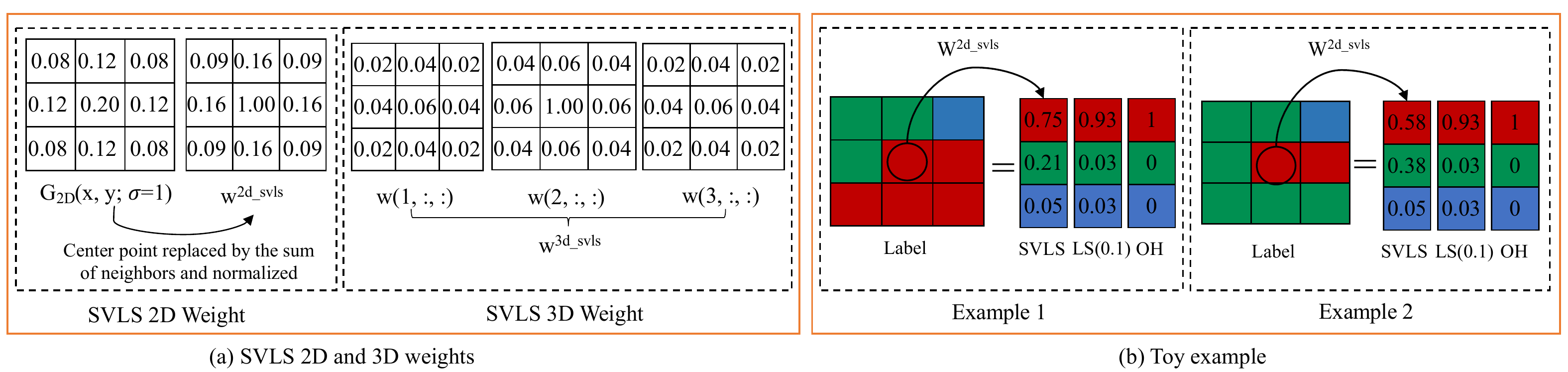}
    \caption{(a) SVLS weights for 2D and 3D. The weights are obtained by re-weighting the center point with the sum of surrounding weights in Gaussian kernel. (b) Two examples of SVLS compared to LS and one-hot labels in two different toy structures. SVLS considers the spatial variation of the pixels and assigns label probabilities accordingly. LS and one-hot ignore the local structure. The center pixel in example 1 has higher probability of being part of the red class than in example 2 due to the spatial context.}
    \label{fig:svls_weights}
\end{figure}

Figure \ref{fig:svls_weights}(a) shows the 2D and 3D versions of SVLS weights. The weight matrix convolves over the one-hot labels to obtain the SVLS soft labels.
\begin{eqnarray}
\label{equ:svls_label}
p_{c(i,j,k)}^{svls} = \frac{1}{\sum w^{svls}} \sum_{x=1}^3 \sum_{y=1}^3 \sum_{z=1}^3 p_{c(i-x,j-y,k-z)} w_{(x,y,z)}^{svls}
\end{eqnarray}
where $p_c^{svls}(i,j,k)$ are soft probabilities after passing proposed weight in the position of $(i,j,k)$ of the one-hot target for class $c$. A single `replicate' padding is required to maintain label consistency around the image. A visual illustration of calculating SVLS weights is presented in Fig.~\ref{fig:weights_labels}(a). The resulting SVLS label probabilities for each class are similar to one-hot within homogeneous areas and thus preserve high confidence in non-ambiguous regions while uncertainty is captured near object boundaries. Fig.~\ref{fig:weights_labels}(b) shows a visualization comparing the label probabilities of one-hot, LS and SVLS for brain tumor example, demonstrating how SVLS captures uncertainty at the tumor boundaries. The resulting SVLS soft labels are then simply plugged into the standard CE loss:
\begin{eqnarray}
CE^{svls} = - \sum_{c=1}^N{p_c^{svls}} \log(\hat{p_c})
\end{eqnarray}

Figure~\ref{fig:svls_weights}(b) illustrates two examples of SVLS labels over LS and one-hot labels in different toy structures. LS and one-hot ignore the local structure where SVLS considers the spatial variation and assigns higher probability to the center pixel in example 1 to be part of the red class based on the spatial context.

\subsection{Multi-rater SVLS}

In cases where where multiple expert annotations are available per image, we introduce a simple extension to SVLS which aggregates uncertainty information across annotations (mSVLS). Here we apply SVLS to each annotation independently before averaging the resulting soft label probabilities. This captures the uncertainty within each expert segmentation and also the variation across the different raters. If there are $D$ number of experts and $p_{c(j)}^{svls}$ is the SVLS target probability for the $j^th$ rater, and $D_c$ number of raters that annotate it as a class then mSVLS probabilities are obtained by simple averaging:
\begin{eqnarray}
\overline p_c^{svls} = \frac{1}{D} \sum_{j=1}^{D_c}{p_{c(j)}^{svls}}
\end{eqnarray}

\begin{figure}[!t]
    \centering
    \includegraphics[width=.9\textwidth]{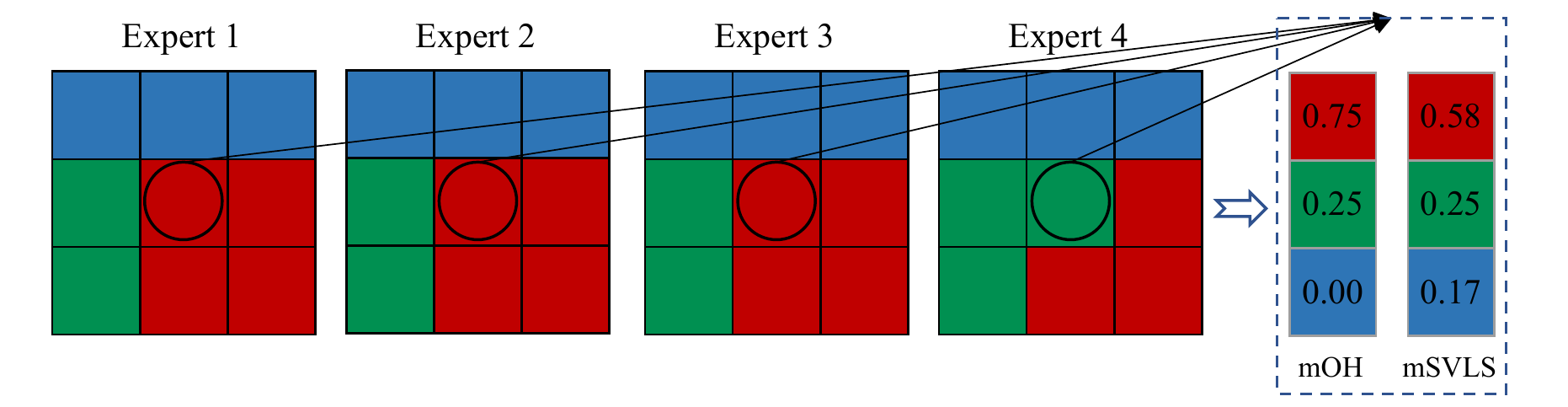}
    \caption{ A toy example showing how to aggregate SVLS soft labels over multiple expert annotations. We apply SVLS to each expert annotation and then average the soft labels to capture inter-rater variability (mSVLS). This is compared to a pixel-wise, independent aggregation of multi-rater one-hot (mOH) labels which would assign zero probability to the blue class despite its presence in the immediate neighborhood.}
    \label{fig:method_msvls}
\end{figure}

Figure \ref{fig:method_msvls} illustrates and compares mSVLS with averaging the one-hot encodings of expert annotations (mOH). The toy example demonstrates how important information about nearby class labels is lost in mOH soft labels, while mSVLS is able to capture the uncertainty across classes and experts.

\section{Experiments}

\subsection{Datasets}
For experimental validation of SVLS, we use three multi-class segmentation datasets including BraTS 2019, KiTS 2019, ProstateX plus one multi-rater dataset, LIDC-IDRI, for binary segmentation. The datasets are described in more detail in the following.

\subsubsection{BraTS 2019}
The BraTS dataset consists of multi-channel MRI for the task of segmenting glioma tumors. There are 335 cases with corresponding annotations in the 2019 training set \cite{bakas2017advancing, bakas2017segmentation, bakas2018identifying}. Each case consists of four MR sequences including FLAIR, T1, T1-contrast, and T2 with image sizes of 155 x 240 x 240. The reference annotations include the necrotic and non-enhancing core, edema and enhancing tumor. All scans are provided pre-processed with skull-stripping, co-registeration to a fixed template, and resampling to 1~mm$^3$ isotropic voxel resolution. We conduct all experiments by splitting the dataset into 269/66 for training and testing. The images are center cropped to size 128 x 192 x 192.
 
\subsubsection{KiTS 2019}
The kidney tumor segmentation challenge (KiTS) \cite{heller2019kits19} consists of 210 CT scans with varying resolutions and image sizes. We resample all cases to a common resolution of 3.22 x 1.62 x 1.62 mm and center crop to image size 80 x 160 x 160. We split the dataset into 168/42 for training and testing. The annotations include the kidney and the tumor  regions.

\subsubsection{ProstateX}
The SPIE-AAPM-NCI ProstateX challenge dataset \cite{litjens2014computer} consists of 98 T2w MRI scans and corresponding annotations for four anatomical prostate zones including the peripheral zone (PZ), transition zone (TZ), distal prostatic urethra (DPU), anterior fibromuscular stroma (AFS). All image volumes are resampled into 3 x 0.5 x 0.5 mm and center cropped to image size 24 x 320 x 320. The dataset is split into 78/20 for training and testing.

\subsubsection{LIDC-IDRI}
The LIDC-IDRI \cite{ armato10data} is a lung CT dataset for nodule segmentation. There are in total 1018 CT scans from which 887 scans contain 4 sets of annotations generated by 12 radiologists. We choose 669 cases for training and the remaining 218 cases for testing. Each CT volume is resampled to 1 mm$^3$ voxel spacing and image sizes of 90 x 90 x 90 centered around the nodule annotation. The PyLIDC library\footnote{https://pylidc.github.io/} is used to perform all pre-processing steps.

\subsection{Implementation Details \& Baselines}
We follow the common pre-processing and data augmentation techniques for all datasets and experiments. The images are center cropped and randomly rotated during data augmentation. A state-of-the-art 3D UNet \cite{cciccek20163d} architecture is used to conduct all experiments with a baseline PyTorch implementation adopted from a publicly available repository\footnote{https://github.com/wolny/pytorch-3dunet}. We use Adam as the optimization method with a learning rate and weight decay set to \num{1e-4}. For comparison, we train models on one-hot target labels and LS soft labels with three different values for the $\alpha$ parameter, namely 0.1, 0.2 and 0.3 (cf. Eq.~(\ref{eq:LS})).

\section{Results} 
 
\subsection{Evaluation Metrics}
We evaluate the segmentation accuracy using the Dice Similarity Coefficient (DSC) and Surface DSC (SD) \cite{nikolov2018deep}. DSC ($= \frac{2\mid T \cap P\mid} {\mid T \mid + \mid P \mid}$) quantifies the overlap between the target reference segmentation $T$ and predicted output $P$. SD assesses the overlap of object boundaries under a specified tolerance level. To evaluate model calibration, we calculate the Expected Calibration Error (ECE) \cite{ guo2017calibration}, Thresholded Adaptive Calibration Error (TACE) \cite{nixon2019measuring} (with a threshold of $10^{-3}$) and plot the reliability diagrams \cite{niculescu2005predicting}.
 
\begin{table*}[!t]
\centering
\caption{Segmentation results for models trained with one-hot hard labels and LS and SVLS soft labels. Dice similarity coefficient (DSC), Surface DSC (SD), Expected Calibration Error (ECE), and Thresholded Adaptive Calibration Error (TACE) metrics are used to evaluate the segmentation accuracy and model calibration. Top two models with highest performance are highlighted as bold and best model is also underlined.}

\scalebox{.85}{\begin{tabular}{|l|c|c|c|c|c|c|c|c|c|c|c|c|c|c|c|c|c|l|} \hline
 &  &  & One-hot & LS (0.1) & LS (0.2) & LS (0.3) & SVLS \\ \hline
\multirow{7}{*}{\rotatebox[origin=c]{90}{BraTS 2019}} & \multirow{3}{*}{DSC} & WT &0.892	$\pm$ 0.074	&0.893	$\pm$ 0.070	&\underline{\textbf{0.896	$\pm$ 0.064}}	&0.893 $\pm$ 0.068	&\textbf{0.894	$\pm$ 0.065}	\\ \cline{3-8}
	 &  & ET &0.814	$\pm$ 0.151	& \underline{\textbf{0.822	$\pm$ 0.154}}	&\textbf{0.817	$\pm$ 0.147}	&0.801	$\pm$ 0.186	&0.816	$\pm$ 0.134\\ \cline{3-8}
	 &  & TC &0.862	$\pm$ 0.126	&\underline{\textbf{0.873	$\pm$ 0.130}}	& \textbf{0.863 	$\pm$ 0.13}1	& 0.858	$\pm$ 0.163	&0.862	$\pm$ 0.119 \\ \cline{2-8}
	 & \multirow{3}{*}{SD} & WT &0.894	$\pm$ 0.074	&0.913	$\pm$ 0.119	&0.912	$\pm$ 0.120	&\textbf{0.92 $\pm$ 0.113} &\underline{\textbf{0.940	$\pm$ 0.086}}\\ \cline{3-8}
	 &  & ET &0.930	$\pm$ 0.162	&\textbf{0.950	$\pm$ 0.121}	&0.943	$\pm$ 0.143	&0.928	$\pm$ 0.159	&\underline{\textbf{0.952	$\pm$ 0.107}}\\ \cline{3-8}
	 &  & TC &0.862	$\pm$ 0.179	&\underline{\textbf{0.891	$\pm$ 0.168}}	&\textbf{0.881	$\pm$ 0.179}	&0.872	$\pm$ 0.190	&0.880	$\pm$ 0.163\\ \cline{2-8}
	 & \multicolumn{2}{c|}{{ECE / TACE}}  &0.073 / 0.0041	&\textbf{0.065 / \underline{0.0025}}	&0.096 / 0.0049	&0.146 / 0.0073	&\textbf{\underline{0.063} / 0.0040}\\ \hline
	 
\multirow{7}{*}{\rotatebox[origin=c]{90}{KiTS 2019}} & \multirow{3}{*}{DSC} & Kidney &\textbf{0.924	$\pm$ 0.048}	& \underline{\textbf{0.924	$\pm$ 0.040}}	&0.917	$\pm$ 0.052	&0.922	$\pm$ 0.045	&0.915	$\pm$ 0.044\\ \cline{3-8}
	 &  & Tumor &0.472	$\pm$ 0.395	&\textbf{0.496	$\pm$ 0.376}	&0.461	$\pm$ 0.382	&0.482	$\pm$ 0.382	&\underline{\textbf{0.491	$\pm$ 0.382}}\\ \cline{3-8}
	 &  & Comp &0.698	$\pm$ 0.221	&\textbf{0.710	$\pm$ 0.208}	&0.689	$\pm$ 0.217	&0.702	$\pm$ 0.213	&\underline{\textbf{0.703	$\pm$ 0.213}}\\ \cline{2-8}
	 & \multirow{3}{*}{SD} & Kidney &0.962	$\pm$ 0.077	&0.960	$\pm$ 0.065	&0.943	$\pm$ 0.098	&\underline{\textbf{0.967	$\pm$ 0.076}}	&\textbf{0.963	$\pm$ 0.076}\\ \cline{3-8}
	 &  & Tumor &0.608	$\pm$ 0.393	&\underline{\textbf{0.637	$\pm$ 0.353}} &0.545 $\pm$ 0.38	&0.617	$\pm$ 0.399	&\textbf{0.634	$\pm$ 0.399}\\ \cline{3-8}
	 &  & Comp &0.785	$\pm$ 0.235	&\textbf{0.798	$\pm$ 0.209}	&0.744	$\pm$ 0.239	&0.792	$\pm$ 0.238	&\underline{\textbf{0.799	$\pm$ 0.238}}\\ \cline{2-8}
	 & \multicolumn{2}{c|}{{ECE / TACE}}  &0.052 / 0.0043	&\textbf{0.042 / 0.0032}	&0.085 / 0.0063	&0.134 / 0.0095	&\underline{\textbf{0.036 / 0.0031}}\\ \hline
	 
\multirow{9}{*}{\rotatebox[origin=c]{90}{ProstateX}} & \multirow{4}{*}{DSC} & PZ &\underline{\textbf{0.716	$\pm$ 0.087}}	&0.684	$\pm$ 0.097	&0.694	$\pm$ 0.098	&\textbf{0.709	$\pm$ 0.094}	& 0.706	$\pm$ 0.082\\ \cline{3-8}
	 &  & TZ &0.850	$\pm$ 0.062	&0.843	$\pm$ 0.071	&0.850	$\pm$ 0.068	&\textbf{0.850	$\pm$ 0.062}	&\underline{\textbf{0.852	$\pm$ 0.056}}\\ \cline{3-8}
	 &  & DPU &\textbf{0.591 $\pm$ 0.169}	&0.589	$\pm$ 0.191	&0.579	$\pm$ 0.181	&0.577	$\pm$ 0.208	&\underline{\textbf{0.621	$\pm$ 0.163}}\\ \cline{3-8}
	 &  & AFS &\underline{\textbf{0.472	$\pm$ 0.110}}	&0.462	$\pm$ 0.097	&0.426	$\pm$ 0.144	&0.466	$\pm$ 0.160	&\textbf{0.467	$\pm$ 0.133}\\ \cline{2-8}
	 & \multirow{4}{*}{SD} & PZ &0.855	$\pm$ 0.106	&0.840	$\pm$ 0.121	&0.828	$\pm$ 0.112	&\textbf{0.858	$\pm$ 0.098} &\underline{\textbf{0.877	$\pm$ 0.083}}\\ \cline{3-8}
	 &  & TZ &\underline{\textbf{0.948	$\pm$ 0.05}}	&0.920	$\pm$ 0.063	&0.916	$\pm$ 0.075	&\underline{0.947	$\pm$ 0.031} &0.935	$\pm$ 0.041\\ \cline{3-8}
	 &  & DPU &0.767	$\pm$ 0.170	&0.792	$\pm$ 0.182	&0.737	$\pm$ 0.188	&\underline{\textbf{0.808	$\pm$ 0.205}}	&\textbf{0.796	$\pm$ 0.192}\\ \cline{3-8}
	 &  & AFS &0.918	$\pm$ 0.099	&0.888	$\pm$ 0.102	&\textbf{0.930	$\pm$ 0.069} &0.900	$\pm$ 0.116	&\underline{\textbf{0.930	$\pm$ 0.066}}\\ \cline{2-8}
	 & \multicolumn{2}{c|}{{ECE / TACE}}  &0.105 / 0.0025	&\textbf{\underline{0.042} / 0.0019}	&\textbf{0.061} / 0.0038	&0.106 / 0.0057	&0.091 / \underline{\textbf{0.0012}}\\ \hline
\end{tabular}}
\label{table:svls_segmentation}
\end{table*}

\subsection{Multi-Class Image Segmentation}
\begin{figure*}[!htbp]
    \centering
    \includegraphics[width=1.0\textwidth]{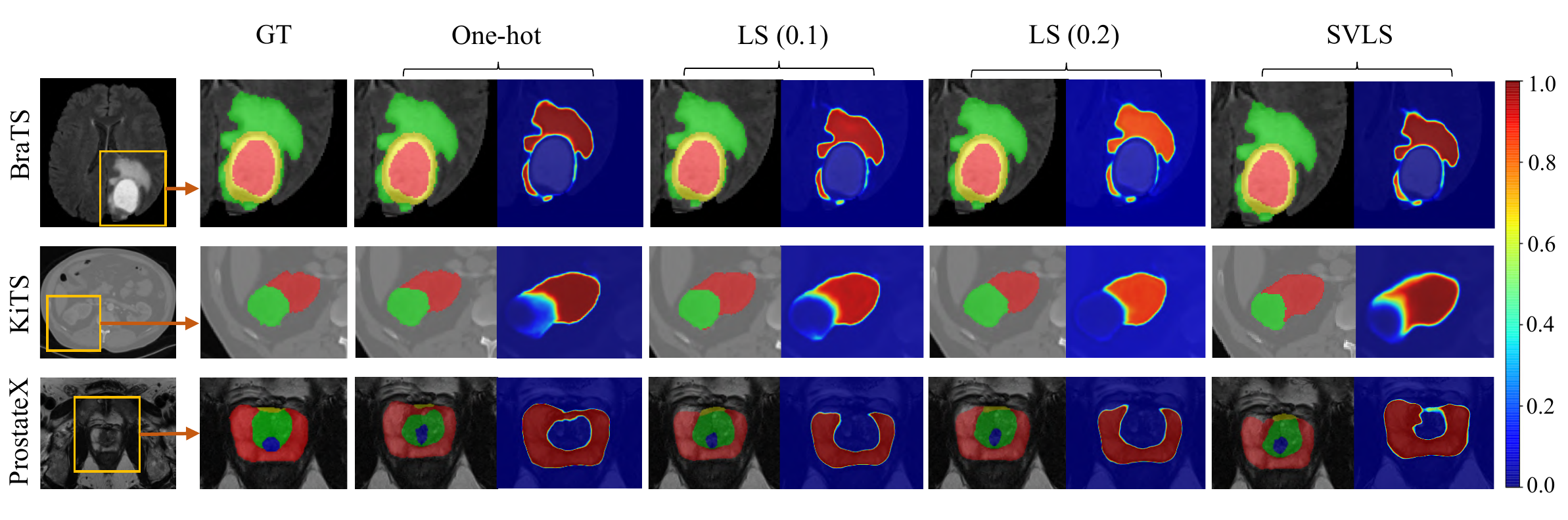}
    \caption{Predicted segmentations with probability maps for BraTS, KiTS, and ProstateX for one-hot, LS and SVLS. We observe reasonable boundary uncertainty for SVLS while preserving high confidence within non-ambiguous regions.}
    \label{fig:svls_probability}
\end{figure*}

\begin{figure*}[!t]
    \centering
    \includegraphics[width=1\textwidth]{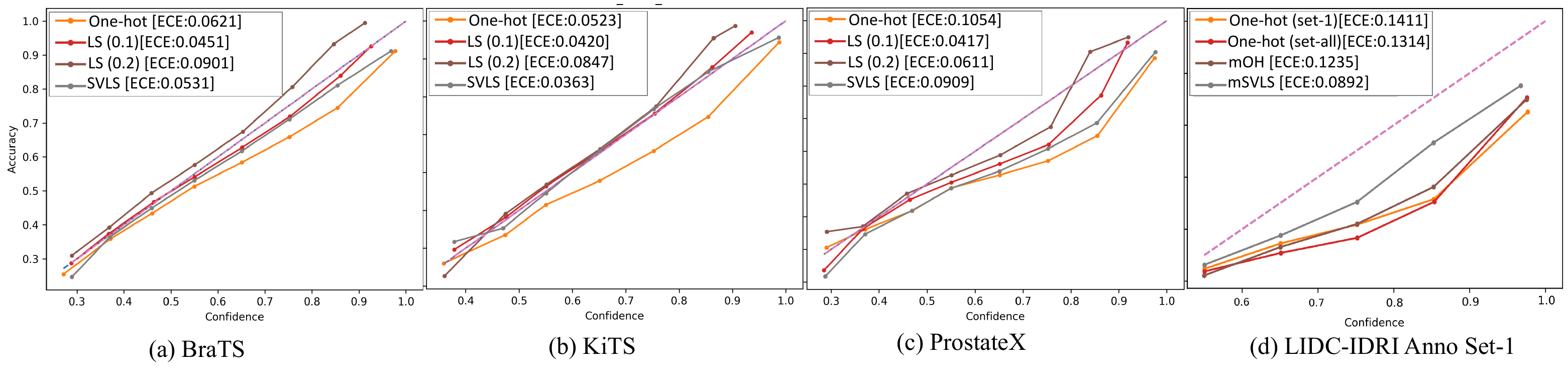}
    \caption{(a-c) Reliability diagrams showing calibration between confidence and accuracy for different models. LS and SVLS have lower expected calibration error (ECE). (d) Reliability diagram on the LIDC-IDRI dataset where multiple expert annotations are available. We show one-hot on annotation set-1, all, multi-rater one-hot (mOH) and mSVLS where the latter shows significantly improved calibration compared to others.}
    \label{fig:svls_msvls_ece}
\end{figure*}

We conduct experiments comparing a state-of-the-art neural network for multi-class image segmentation trained with and without LS and SVLS using CE loss as the training objective. The results are summarized in Table~\ref{table:svls_segmentation}, Fig.~\ref{fig:svls_msvls_ece}(a-c) and Fig.~\ref{fig:svls_probability}. We observe a significant improvement of model calibration in terms of ECE, TACE, and confidence frequency plots as well as better segmentation accuracy for LS and SVLS over the baseline model trained with one-hot target labels. The improvement in the boundary regions is highlighted by the SD metric. In Table~\ref{table:svls_segmentation}, the SD score is calculated with a tolerance level of 2 mm, and SVLS shows 2-4\% improvement in accuracy over the baseline using one-hot labels across all three datasets, BraTS 2019, KiTS 2019 and ProstateX. SVLS and LS both improve model calibration seen by improved ECE scores with LS obtaining best ECE on BraTS and ProstateX. SVLS shows best results for model calibration under the TACE metric over all datasets.

The reliability diagrams also indicate better model calibration for SVLS as shown in Fig.~\ref{fig:svls_msvls_ece}(a-c). Optimal calibration between confidence and accuracy would be obtained when aligning with the dashed line (bottom row). For the baseline, we observe that the accuracy is always lower than the confidence, indicating that the model trained on one-hot labels yields overconfident predictions. Fig.~\ref{fig:svls_probability} visualizes the prediction and corresponding probability maps. The prediction probability of SVLS shows reasonable uncertainty in boundary regions between classes and anatomical regions.

\begin{table*}[!ht]
\centering
\caption{Quantitative comparison for LIDC-IDRI with multiple expert annotations. Dice Similarity Coefficient (DSC), Surface DSC (SD), Expected Calibration Error (ECE), and Thresholded Adaptive Calibration Error (TACE) are calculated to compare the performance of one-hot, LS, and mSVLS. Top two models with highest performance are highlighted as bold and best model is also underlined.}
\scalebox{.77}{\begin{tabular}{|l|c|c|c|c|c|c|c|c|c|c|c|c|c|c|c|c|c|l|} \hline
  &  & One-hot (Set 1) & One-hot (Set 2) & One-hot (Set 3) & One-hot (ALL) & mOH & mSVLS \\ \hline
\multirow{4}{*}{\rotatebox[origin=c]{90}{DSC}} & Set 1 &0.687 $\pm$ 0.181	&0.676 $\pm$ 0.192	&0.669 $\pm$ 0.206	&\textbf{0.691 $\pm$ 0.175}	&0.690 $\pm$ 0.180	&\underline{\textbf{0.692 $\pm$ 0.171}} \\ \cline{2-8}
& Set 2 &0.674 $\pm$ 0.185	&0.652 $\pm$ 0.195	&0.651 $\pm$ 0.218	&\underline{\textbf{0.683 $\pm$ 0.173}}	&0.675 $\pm$ 0.188	&\textbf{0.680 $\pm$ 0.183} \\ \cline{2-8}
& Set 3 &0.684 $\pm$ 0.179	&0.670 $\pm$ 0.194	&0.666 $\pm$ 0.205	&0.686 $\pm$ 0.172	&\textbf{0.687 $\pm$ 0.180}	&\underline{\textbf{0.688 $\pm$ 0.181}} \\ \cline{2-8}
& Set 4 &0.679 $\pm$ 0.182	&0.655 $\pm$ 0.197	&0.662 $\pm$ 0.213	&\underline{\textbf{0.694 $\pm$ 0.172}}	&0.682 $\pm$ 0.183	&\textbf{0.686 $\pm$ 0.174} \\ \cline{1-8}
	 
\multirow{4}{*}{\rotatebox[origin=c]{90}{SD}} & Set 1 &0.914 $\pm$ 0.171	&0.892 $\pm$ 0.177	&0.910 $\pm$ 0.192	&\textbf{0.936 $\pm$ 0.145}	&0.924 $\pm$ 0.158	&\underline{\textbf{0.941 $\pm$ 0.132}} \\ \cline{2-8}
& Set 2 &0.905 $\pm$ 0.172	&0.880 $\pm$ 0.178	&0.897 $\pm$ 0.202	&\textbf{0.927 $\pm$ 0.150}	&0.915 $\pm$ 0.157	&\underline{\textbf{0.932 $\pm$ 0.139}}	\\ \cline{2-8}
& Set 3 &0.915 $\pm$ 0.171	&0.890 $\pm$ 0.180	&0.905 $\pm$ 0.198	&\textbf{0.935 $\pm$ 0.153}	&0.923 $\pm$ 0.162	&\underline{\textbf{0.936 $\pm$ 0.148}}	\\ \cline{2-8}
& Set 4 &0.914 $\pm$ 0.168	&0.884 $\pm$ 0.182	&0.905 $\pm$ 0.198	&\textbf{0.934 $\pm$ 0.148}	&0.921 $\pm$ 0.156	&\underline{\textbf{0.939 $\pm$ 0.134}}	\\ \cline{1-8}
	 
\multirow{4}{*}{\rotatebox[origin=c]{90}{\scriptsize ECE/TACE}} & Set 1 & 0.141 / 0.0032	& 0.1457 / 0.0035	& 0.1646 / 0.0042	& 0.1314 / 0.0029	& \textbf{0.1234 / \underline{0.0022}}	& \textbf{\underline{0.0892} / 0.0028} \\ \cline{2-8}
& Set 2 & 0.123 / 0.0037	& 0.1304 / 0.0041	& 0.1401 / 0.0044	& 0.1152 / 0.0033	& \textbf{0.0993 / \underline{0.0024}}	& \textbf{\underline{0.0692} / 0.0030}	\\ \cline{2-8}
& Set 3 & 0.1271 / 0.0031	& 0.1341 / 0.0039	& 0.1409 / 0.0035	& 0.1195 / 0.0030	& \textbf{0.1053 / \underline{0.0025}}	& \textbf{\underline{0.0748} / 0.0028}	\\ \cline{2-8}
& Set 4 & 0.1098 / 0.0029	& 0.1212 / 0.0039	& 0.1301 / 0.0033	& 0.1032 / 0.0025	& \textbf{0.0948 / \underline{0.0022}}	& \textbf{\underline{0.0653} / 0.0025}	\\ \hline
	 
\end{tabular}}
\label{table:wl_svls_segmentation}
\end{table*}

\subsection{Multiple Expert Annotations}
To evaluate the performance of mSVLS, we run experiments by training models on different sets of expert annotations. The compared models are trained with individual expert annotations, all pairs of annotations, multi-rater one-hot labels (mOH) and mSVLS, evaluated against each available expert annotation, separately. Table \ref{table:wl_svls_segmentation} shows the results for the metrics DSC, SD, ECE and TACE. mSVLS shows overall best performance in most cases. 
The model calibration is illustrated in Fig. \ref{fig:svls_msvls_ece}(d) in terms of confidence frequency distribution (top) and ECE (bottom). In the reliability diagram, the curve for mSVLS is closer to optimal model calibration (dashed line).

\begin{figure*}[!htbp]
    \centering
    \includegraphics[width=1\textwidth]{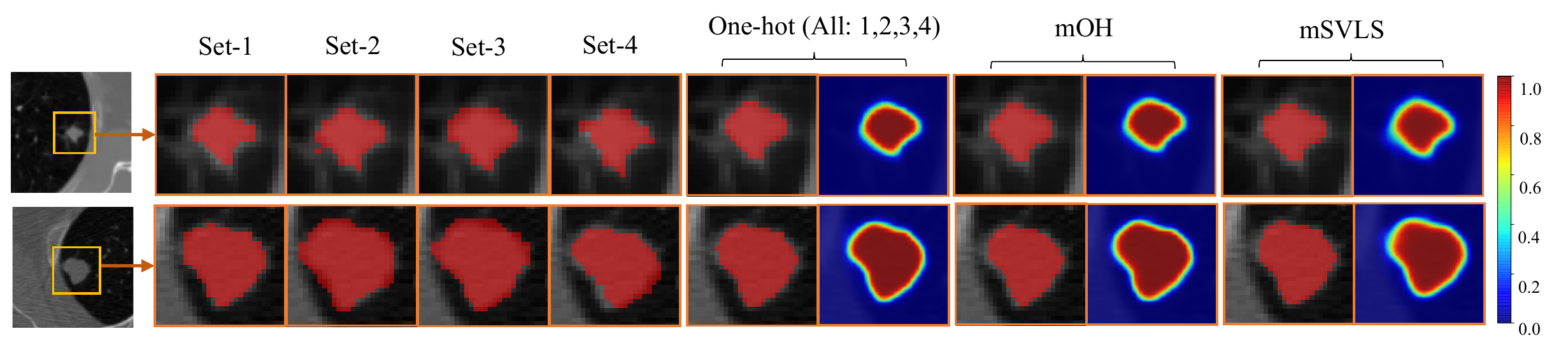}
    \caption{Comparison of all four sets of annotations and corresponding CT scan (left five columns) and model predictions and probability maps for one-hot encoding (trained with the fusion of all sets), multi-rater one-hot (mOH), and multi-rater SVLS (mSVLS) on LIDC-IDRI dataset (right three pairs of columns, respectively). mSVLS shows reasonable boundary uncertainty and high confidence within the structure compared to the one-hot and mOH showing uniform confidence levels across structure boundaries.}
    \label{fig:wl_svls_probability}
\end{figure*}

Figure~\ref{fig:wl_svls_probability} indicates the qualitative segmentation performance and corresponding prediction probability map. mSVLS produces reasonable uncertainty at the boundary while being confident in non-ambiguous regions. The one-hot and mOH baselines show uniform confidence levels across the structure boundaries, ignoring spatial variability among the expert annotations.
 
\section{Discussion}
We propose a novel, simple yet effective approach for capturing uncertainty from expert annotations which also improves model calibration in semantic segmentation. Promising results are demonstrated on four different datasets, including CT and MRI data, and multi-class and binary segmentation tasks. We also demonstrate how SVLS can capture expert variation when multiple annotations are available. SVLS is straightforward to integrate into any existing segmentation approach, as it only requires a pre-processing on the one-hot encodings before training. Here, we trained all models with a cross-entropy loss, but other losses such as boundary losses \cite{lee2020structure} can be considered. We like to highlight that SVLS is complementary to other works on estimating uncertainty in image segmentation, and can be considered in combination, e.g., with approaches that incorporate uncertainty as part of the output predictions \cite{baumgartner2019phiseg,monteiro2020stochastic}.

An interesting aspect that should be explored further is the effect of label smoothing on the separation of the learned latent representations for different classes. In \cite{muller2019does}, it was shown that traditional LS has a similar effect as contrastive learning pushing the latent representations away from each other. This may be interesting in the context of domain adaptation and robustness on variations in the input data, for example, when images at test time come from a different distribution (e.g., different scanner). The effect of SVLS on the learned latent representations and how it impacts generalization is part of future work. 

\section*{Acknowledgements}
This research has received funding from the European Research Council (ERC) under the European Union's Horizon 2020 research and innovation programme (grant agreement No 757173, project MIRA).

\bibliography{mybib}{}
\bibliographystyle{splncs04}
\end{document}